\pdfoutput=1
\documentclass[11pt]{article}

\usepackage[final]{coling}

\usepackage{xcolor}
\usepackage{times}
\usepackage{latexsym}
\usepackage{algorithm}
\usepackage{algorithmic}
\usepackage{algorithm}
\usepackage{algorithmic}
\usepackage{amsmath} 
\usepackage{makecell} % 线条宽度
\usepackage{arydshln}
\usepackage{multirow} % 导入 multirow 宏包
\usepackage[T1]{fontenc}
\usepackage[utf8]{inputenc}

\usepackage{microtype}

\usepackage{inconsolata}

\usepackage{graphicx}

\title{Low-Resource Fast Text Classification Based on Intra-Class and Inter-Class Distance Calculation}

\newcommand*\samethanks[1][\value{footnote}]{\footnotemark[#1]}
\newcommand*\commonmark{\textsuperscript{†}}
\author{
 \textbf{Yanxu Mao\textsuperscript{1}\thanks{Authors contribute equally}},
 \textbf{Peipei Liu\textsuperscript{2,3}\samethanks\thanks{Corresponding author}},
 \textbf{Tiehan Cui\textsuperscript{1}},
 \textbf{Congying Liu\textsuperscript{3}},
 \textbf{Datao You\textsuperscript{1}\commonmark}
\\
 \textsuperscript{1}School of Software, Henan University, China\\
 \textsuperscript{2}Institute of Information Engineering, Chinese Academy of Sciences, China\\
 \textsuperscript{3}University of Chinese Academy of Sciences, China
\\
 \small{
   \textbf{Correspondence:} \href{mailto:email@domain}{peipliu@yeah.net}
 }
}

\begin{document}
\maketitle
\begin{abstract}
In recent years, text classification methods based on neural networks and pre-trained models have gained increasing attention and demonstrated excellent performance. However, these methods still have some limitations in practical applications: (1) They typically focus only on the matching similarity between sentences. However, there exists implicit high-value information both within sentences of the same class and across different classes, which is very crucial for classification tasks. (2) Existing methods such as pre-trained language models and graph-based approaches often consume substantial memory for training and text-graph construction. (3) Although some low-resource methods can achieve good performance, they often suffer from excessively long processing times. To address these challenges, we propose a low-resource and fast text classification model called \textbf{LFTC}. Our approach begins by constructing a compressor list for each class to fully mine the regularity information within intra-class data. We then remove redundant information irrelevant to the target classification to reduce processing time. Finally, we compute the similarity distance between text pairs for classification. We evaluate \textbf{LFTC} on 9 publicly available benchmark datasets, and the results demonstrate significant improvements in performance and processing time, especially under limited computational and data resources, highlighting its superior advantages. 
\end{abstract}

\section{Introduction}
Text classification aims to categorize natural language texts into predefined classes \cite{minaee2021deep}, and it is widely used in various fields such as sentiment analysis, topic classification \cite{yang2016hierarchical}, and news classification. Currently, deep learning based methods represented by neural networks dominate in text classification tasks \cite{lin2021bertgcn, li2022survey}. Existing methods can be divided into two categories \cite{ding2020more, lin2021bertgcn, jiang2023low}: transductive learning, represented by graph neural networks, and inductive learning, represented by recurrent neural networks and convolutional neural networks. However, transductive learning methods require access to the test dataset during the training phase \cite{li2021textgtl}, which means that when encountering new text data, the existing model needs to be retrained. This limitation reduces the practical applicability of these methods. Therefore, this paper focuses on inductive learning methods for text classification.

Existing text classification models \cite{lin2021bertgcn,devlin2019bert} typically rely on large amounts of labeled data and high-performance computing resources to achieve their superior performance. While these models excel at handling large-scale data, their application in low-resource settings (e.g., when labeled data is scarce or computational power is limited) is constrained \cite{zhao2022improving}. In cases of few-shot learning, these neural network-based models exhibit a certain degree of robustness. However, their limited feature representation often falls short of meeting practical application needs. Recently, \citet{jiang2023low} proposed a classification method based on a single compressor, which to some extent alleviates the issues of data scarcity and limited computational resources. \citet{wen2023augmenting} employed graph-based pre-training and prompts to enhance low-resource text classification. These methods not only achieve efficient classification results on limited datasets but also significantly reduce model complexity and computational costs.

Despite previous research achieving breakthrough results, real-world applications may face significant limitations in terms of speed and resource requirements \cite{liu2024improved, ding2020more}. These methods have the following three main limitations: (1) Existing deep learning methods mainly focus on simple pairwise sentence matching within texts (i.e., inter-sentence relationships). However, in natural language texts, the interactions between sentence pairs are not merely binary as a single sentence may have close connections with many other sentences within the texts. This necessitates greater attention to intra-class regularities and inter-class differences to more effectively complete the classification task. (2) Current methods often have high computational resource requirements. Approaches based on pre-trained language models \cite{lin2021bertgcn} and graph-based methods \cite{wang2022induct} can result in significant memory consumption during training and text word graph construction. (3) Some low-resource methods (i.e., methods with limited data and computational resources \cite{ding2020more}) still require significant time consumption in pursuit of higher efficiency. The time costs associated with these methods limit their practicality in real-time applications. Therefore, although these methods offer certain solutions in resource-constrained environments, there is still a need to balance efficiency with time consumption in practical applications to enhance their real-world value.

To mitigate the limitations of existing methods, we propose an efficient and rapid text classification approach that does not require a pre-training process and is parameter-free. This approach achieves rapid processing in environments with constrained computational and data resources by optimizing data handling and classification strategies. Specifically, by employing innovative and efficient data structures, we significantly reduce time complexity and computational overhead. Additionally, this approach is adaptable to various text classification tasks without relying on large-scale pre-trained models, thus reducing the complexity of implementation and maintenance. Experiments on multiple benchmark datasets demonstrate that this method enhances classification accuracy while significantly improving processing speed and resource utilization. Our main contributions can be summarized as follows:
\begin{itemize}
\item We propose \textbf{LFTC} (\textbf{L}ow-Resource \textbf{F}ast \textbf{T}ext \textbf{C}lassification), which utilizes a text compression method to calculate the compression distance of global and local text information. The approach fully leverages multiple inter-class and intra-class correlations to achieve text classification tasks.
\item \textbf{LFTC} is a lightweight model, especially suitable for scenarios with scarce labeled data and limited computational resources. The model effectively eliminates redundant data irrelevant to predictions, thereby completing text classification tasks in a relatively short time and demonstrating high practicality in real-world applications.
\item We conduct extensive experiments on nine benchmark datasets, and our method achieves SOTA scores on multiple datasets among non-pretrained models. The method also significantly outperforms others in few-shot experiments, demonstrating the model's superiority.
\end{itemize}

\section{Related Work}
Our research is closely related to text classification methods based on data compression, low-resource, and deep learning. Therefore, we have provided a brief overview of these three methods.
\subsection{Text Classification Based on Data Compression}
This is a relatively uncommon approach, where methods calculate the similarity score between texts based on the compression distance derived from a specific compression technique, thereby accomplishing text classification tasks \cite{keogh2004towards}. Initially, \citet{benedetto2002language} proposed a text classification method that combined entropy estimation and the open Gzip compressor with text similarity measurement. Subsequently, \citet{coutinho2015text} introduced a text classification method based on information-theoretic dissimilarity measures, mapping texts into a feature space defined by these measures to represent dissimilarity. Later, \citet{kasturi2022text} presented a language-agnostic technique called Zest, which further improved the performance of text classification tasks by simplifying configuration and enhancing text representation, thus avoiding meticulous feature extraction and large models. Recently, \citet{jiang2023low} proposed a single-compressor model called \textbf{gzip}, which combines the open Gzip compressor and a classifier for text classification tasks without any training parameters. However, although existing methods can provide excellent performance, they often require longer processing times. 
% Our new method not only improves performance but also reduces time consumption, achieving dual optimization.

\subsection{Low Resource Text Classification}
% Low-resource text classification refers to the task of classifying text when there is very limited labeled data (text samples with labeled categories) available \cite{niyongabo2020kinnews},
Low-resource text classification refers to the task of classifying text when labeled data (i.e., text samples with classification labels) is extremely limited, and it can also be considered as having limited computational resources. These situations are quite common in practical applications, as collecting large amounts of labeled data is often time-consuming and computational resources are expensive. \citet{ding2020more} proposed a principled model called Hypergraph Attention Networks, which can achieve greater expressive power with less computational cost, used for text representation learning. In low-resource text classification, the scarcity of labeled data can lead to poor performance in traditional classification models that require large amounts of labeled data for training, and may even result in overfitting issues \cite{hedderich2021survey}. \citet{chen2022contrastnet} introduced a contrastive learning framework called ContrastNet, which addresses the issues of discriminative representation and overfitting in text classification by learning to pull together text representations of the same class and push apart those of different classes.

\subsection{Text Classification Based on Deep Learning}
\citet{zhang2015character,adhikari2019rethinking} utilized CNN convolutional layers to extract local features from text, capturing $n$-gram features for classification. \citet{wang2016attention} proposed an attention-based LSTM that processes text sequence data and captures long-distance dependencies within the text, demonstrating competitive performance in aspect-level text classification. Subsequently, \citet{devlin2019bert} introduced the pre-trained language model BERT, which uses the self-attention mechanism to capture contextual dependencies in text, achieving high performance in text classification. Later, \citet{lin2021bertgcn,sun2023text,liu2024improved} proposed methods that leverage large-scale pre-training on massive raw data and jointly learn representations for both labeled training data and unlabeled test data through label propagation using graph convolutional networks (GCNs). However, these methods typically require substantial computational and data resources, making them challenging to apply effectively in low-resource environments. 
% Our proposed \textbf{LFTC} significantly alleviates the consumption of resources and time.

Compared to the existing methods mentioned above, our method not only improves performance but also reduces time consumption, achieving dual optimization.

\section{Method}
% In this section, we describe the process of constructing the compressor list and provide a detailed explanation of the proposed \textbf{LFTC} model. The model mainly consists of two parts: the Multi Compressor Classification Layer and the Centralized Reasoning Layer. To aid in understanding, we present the overall framework of the \textbf{LFTC} in Figure~\ref{lftc_model}, and the pseudocode for the model execution process is shown in Algorithm~\ref{alg:lftc_model}.

\begin{figure}[t]
  \includegraphics[width=\columnwidth]{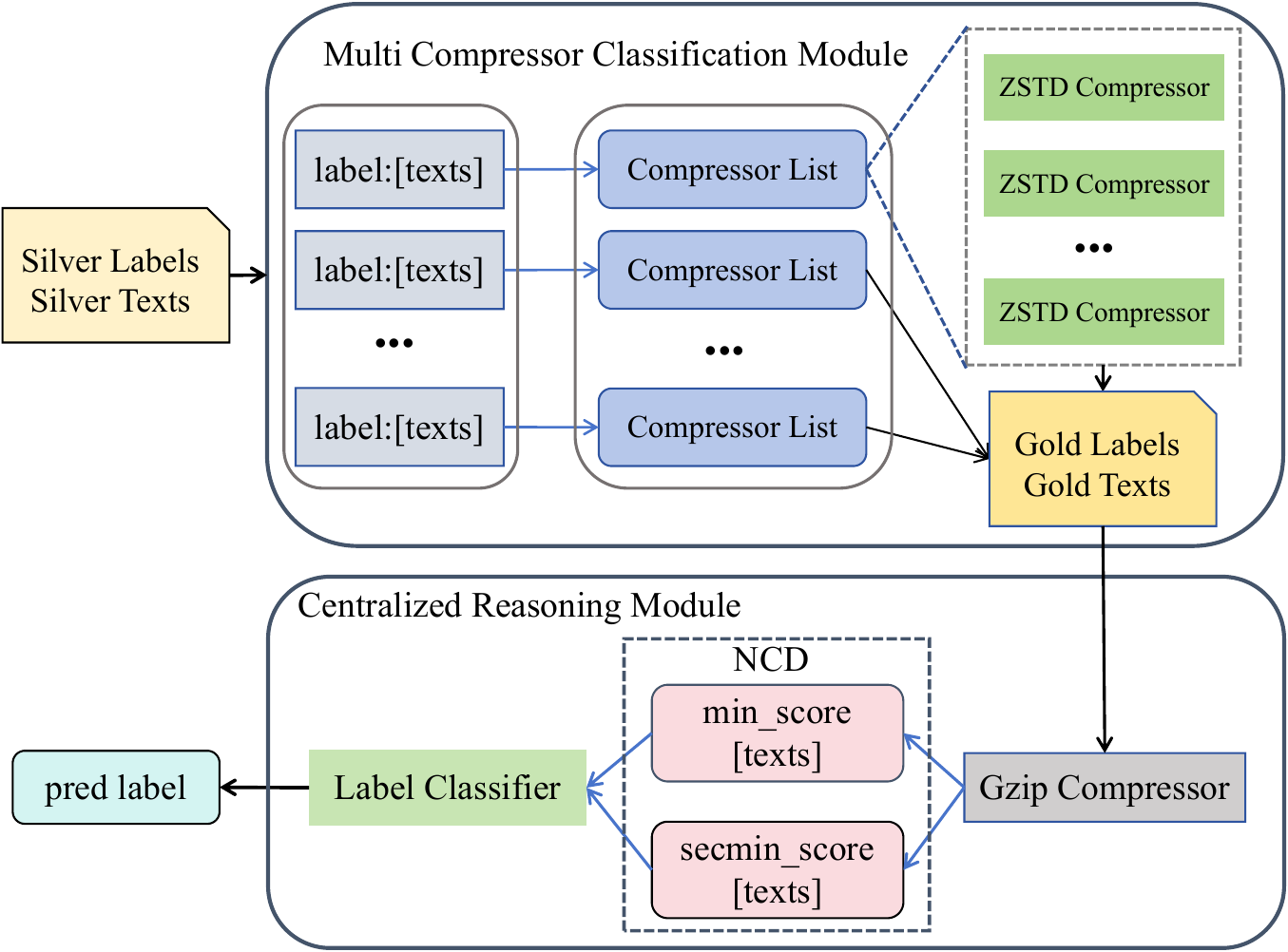}
  \caption{The overall architecture of \textbf{LFTC}.}
  \label{lftc_model}
\end{figure}

Figure~\ref{lftc_model} presents the overall framework of our proposed \textbf{LFTC} model, and Algorithm~\ref{alg:lftc_model} shows the pseudocode corresponding to the model's execution process (Due to space limitations, we have included it in Appendix A). In this section, we first describe the construction process of the compressor list, followed by a detailed explanation of the two execution modules of \textbf{LFTC}: the Multi Compressor Classification Module and the Centralized Reasoning Module.

\subsection{Compressor List Construction}

Text compression algorithms reduce the storage space of text by removing redundant data. The different compression lengths produced by applying the same compression algorithm to different texts can reflect the varying characteristics of the text content \cite{kasturi2022text}. Texts within the same class exhibit more regularity compared to those from different classes \cite{jiang2023low}. 
% Therefore, to better utilize the intra-class information, we extract and concatenate the texts of each class $C_i$ from the training data $D=\{T_1, T_2, \ldots, T_n\}$ as follows: $Ts_{C_i}=\{T_{1_{C_i}}, T_{2_{C_i}}, \ldots, T_{n_{C_i}}\}$, where $Ts_{C_i}$ represents all the texts within the category.
% Therefore, to better utilize the intra-class information, we extract and concatenate the texts of each class $C_i$ from the training data $D=\{T_1, T_2, \ldots, T_m\}$ as follows: 
% $Ts_{C_i}=\{T_{1_{C_i}}, T_{2_{C_i}}, \ldots, T_{l_{C_i}}\}$, where $Ts_{C_i}$ represents all the intra-class texts.
Therefore, to better utilize the intra-class information, we find and concatenate the all texts belonging to each class $C_i$ from the training data with $m$ texts $\{T_1, T_2, \ldots, T_m\}$ : 
$Ts_{C_i}=\{T_{1_{C_i}}, T_{2_{C_i}}, \ldots, T_{l_{C_i}}\}$, where $Ts_{C_i}$ represents all the intra-class texts of $C_i$ class.

We divide the data of each class into $N_{C_i}$ segments based on the given step size $S$:
\begin{align}
\begin{aligned}
N_{C_i} = \left\lceil \frac{len(Ts_{C_i})}{|S|} \right\rceil
\end{aligned}
\end{align}

For each segment, a compressor is constructed:
\begin{align}
\begin{aligned}
Z_{C_i, N_{C_i}} = Zstd\left(Ts_{C_i}\left[x \cdot S : (x+1) \cdot S\right]\right)
\end{aligned}
\end{align}
where $x$ is the segment index. $Zstd$ is a high-speed lossless data compression algorithm \cite{chen2021fpga}. 

Suppose the text $T = \{L_1, L_2, \ldots, L_n\}$, where $L_n$ is the $n$-th substring of the text. A compression dictionary $\mathcal{D}$ is constructed, assigning corresponding labels $p$ to the substrings appearing in the text. Finally, we integrate these compressors built based on intra-class regularities of the same text and obtain a set of compressors corresponding to each class $C_i$:
% Finally, we construct a list of compressors based on the intra-class regularities of the text and obtain a set of compressors corresponding to each class $C_i$:
% the compressors trained based on the intrinsic rules of texts with the same labels are saved as a list of compressors, resulting in a set of compressors corresponding to each category:
\begin{align}
\begin{aligned}
Z_{C_i} = \{Z_{C_i, 0}, Z_{C_i, 1}, \ldots, Z_{C_i, N_{C_i}-1}\}.
\end{aligned}
\end{align}

\subsection{Multi Compressor Classification Module}
% We input the data to be classified, $T_{C_i}$, into the constructed compressors. Based on the characteristics of the input data, we use the adaptive algorithm Zstd.
% % \footnote{It is from Facebook: \url{https://github.com/facebook/zstd}.}
% to adjust the compression level to optimize both compression speed and compression ratio. Each text $T_{C_i}$ is processed through all the compressor lists corresponding to the labels. The compressors maintain a sliding window $W$ for storing and searching recently seen strings. This allows us to calculate the maximum matching substring $\text{L}_{\max}$ in the current text $T_{C_i}$:

We input the data to be classified, $T_{C_i}$, into the constructed compressors. Based on the characteristics of the input data, we use Zstd's adaptive algorithm to adjust the compression level for optimizing both compression speed and compression ratio. Each text $T_{C_i}$ is processed through all the compressors corresponding to the each label. The compressors maintain a sliding window $W$, which is used to store and search for recently seen strings. This allows us to compute the longest matching substring $\text{L}_{\max}$ in the current text $T_{C_i}$:
\begin{align}
\begin{aligned}
\scalebox{0.85}{$
\text{L}_{\max} = \max\{\text{L}:T_{C_i}[j: j+W]= \mathcal{D}[k,k+W]\}$}
\end{aligned}
\end{align}
where $W \in [ 1,\text{ $len{(T_{C_i})}$ }]$, ${len}(\cdot)$ calculates the length of the text, $j \in [0,\text{ $len{(T_{C_i})}$ }-\text{ $W$ }]$, $k \in [1,j)$, and \text{L} is a substring of the text $T_{C_i}$.

We replace the repeated strings with the corresponding labels $ p $ from the compression dictionary $\mathcal{D}$ based on the maximum matching substring, thereby reducing the data volume. The replaced string $L_{\text {d}}$ can be represented as:
\begin{align}
\begin{aligned}
L_{\text {d}} = T_{C_i}:\operatorname{Re}(\text { L }_{\max }, p)
\end{aligned}
\end{align}
During the compression process, entropy coding is used to assign shorter codes to high-probability symbols and longer codes to low-probability symbols, further reducing the data size. The memory size obtained by entropy encoding $T_{C_i}$ is:
\vspace{-1ex}
\begin{align}
\begin{aligned} 
\mathcal{L}e = -\sum_{d=1}^n P_{(L_d)} \cdot \log_2{P_{(L_d)}}
\end{aligned}
\end{align}
where $n$ is the number of $T_{C_i}$'s substrings $L_d$, and $P_{(L_d)}$ is the probability of occurrence of substring $L_d$.

% Subsequently, we can calculate the compressed length of the text $T $ based on the lengths obtained from dictionary replacement and entropy coding as follows:
Subsequently, we can calculate the final compression length of the input text $T_{C_i}$ after dictionary replacement and entropy encoding. The value, defined as the final score of compressor $Z_{C_i, N_{C_i}}$ with class $C_i$, can be expressed as follows:
\begin{align}
\begin{aligned}
Score(Z_{C_i, N_{C_i}}) = \mathcal{L}e(T_{C_i}) + O(\mathcal{D})
\end{aligned}
\end{align}
where $\mathcal{L}e(T_{C_i})$ is the actual space occupied by the compressed data. And $O(\mathcal{D})$ represents the additional memory overhead required for using the compression dictionary, which includes the storage overhead of the dictionary itself and other metadata costs associated with the compression process.

Finally, we sum the compression scores for each compressor list $Z_{C_i}$ to obtain the final score for the current text under each label. The shorter the compression length, the lower the score, indicating that the model is more familiar with the text of that category \cite{kasturi2022text, jiang2023low}. We search the Silver data for the two texts $T_p$ and $T_q$ corresponding to the lowest scores, with their category labels denoted as $p$ and $q$, respectively.

\subsection{Centralized Reasoning Module}
To achieve more accurate predictions, we extract the text data with classification labels $p$ and $q$ from the training data for centralized information inference. This approach better utilizes the relevant information between two classes and requires only localized computations, thereby excluding redundant data and significantly improving the model's prediction speed. To search for other text data most similar to the true class of the current text $T_{C_i}$, we remove $T_p$ and $T_q$ while extracting data with class labels $p$ and $q$. The remaining data is used as supporting evidence for focused inference, and we refer to this text data as Gold data.

First, the Gold data undergoes a simple compression process using the Gzip compressor. Second, we use the Normalized Compression Distance (NCD) \cite{cohen2015normalized} to measure the similarity between the prediction text $T$ and the Gold data. It is computed as follows:
\begin{align}
\begin{aligned}
\scalebox{0.95}{$
N C D\left(T_{C_i}, \mathcal{Y}\right) =
\frac{C\left(T_{C_i}\mathcal{Y}\right) - \min \left(C(\mathcal{Y}), C(T_{C_i})\right)}{\max \left(C(\mathcal{Y}), C(T_{C_i})\right)}$}
\end{aligned}
\end{align}
where $C(\cdot)$ represents the compression size, and $\mathcal{Y}=(Ts_{C_p}:Ts_{C_q})$ is the concatenation of the two labeled datasets.

Through the above steps, we can obtain the compression distance between the input text and the Gold data. We use the \textit{KNN} algorithm to classify a data point based on its distance from other points. Given a sample $T_{C_i}$ to be classified, the algorithm identifies the $K$ nearest samples in the Gold data that are most similar to $T_{C_i}$ (i.e., the $K$ nearest neighbors). The class of the sample is then determined by voting or weighting based on the labels of these neighboring samples.
% By calculating the distance between the new data point and all points in the training dataset, we select the $k$ nearest neighbors to determine its classification. 
% We choose the minimum compression distance (i.e., $k_{nn}=1$) as the final predicted label and output it. This approach avoids the tie situations that arise in the method proposed by \cite{jiang2023low} (They select the two smallest distances as the prediction results).

\section{Experiments}

\subsection{Datasets}

To validate the effectiveness of \textbf{LFTC}, we conducted experiments on nine benchmark datasets widely used in text classification tasks. These datasets cover a range of content from technical reports to medical literature, and provide social news from different languages and cultural backgrounds. These characteristics make them ideal for assessing the effectiveness and generalization capability of text classification models. A summary of the statistics on categories, sample sizes, and other details for each dataset is presented in Table~\ref{dataset}, with a detailed description provided below.
(1) \textbf{R8} and \textbf{R52} \cite{joachims1998text} are two Reuters datasets used for news classification.
(2) \textbf{AGnews} \cite{del2005ranking} is sourced from the online academic news search engine comeToMyHead, featuring a moderate amount of data, balanced category distribution, and text content covering multiple domains.
(3) \textbf{Ohsumed} \cite{hersh1994ohsumed} is a medical dataset containing 270 types of medical literature.
(4) \textbf{SogouNews} \cite{zhang2015character} is a Chinese news classification dataset provided by Sogou, including news articles collected from the Sogou News website.
(5) \textbf{20News} \cite{lang1995newsweeder} is a classic English text classification dataset containing posts from 20 different newsgroups.
(6) \textbf{SwahiliNews} \cite{martin2022swahbert} is a dataset for Swahili news classification, while \textbf{kirnews} and \textbf{kinnews} \cite{niyongabo2020kinnews} are datasets for news classification in Kirundi and Kinyarwanda, respectively. These datasets were created to support NLP research for minority languages.

\begin{table}
    \centering
    \resizebox{0.90\columnwidth}{!}{
    \begin{tabular}{llcccccccc}
    \Xhline{1.0pt}
    
    \multirow{1}{*}{Dataset}
    % \hline
    & Train && Test && Class && Word \\
    \hline
    R8  &5.5K &&2.2K&& 8 &&24K\\
    R52 &6.5K &&2.6K&&52&&26K\\
    Ohsumed  &3.4K &&4K&&23&&55K\\
    20News &11K &&7.5K&&20&&277K\\
    AGnews &120K &&7.6K&&4&&128K\\
    kirnews &3.7K &&0.9K&&14&&63K\\
    kinnews &17K &&4.3K&&14&&240K\\
    SwahiliNews &22.2K && 7.3K && 6 &&570K\\
    SogouNews &450K&&60K&&5&&611K\\
    
    \Xhline{1.0pt}
    \end{tabular}}%
    \caption{\label{dataset}
    Summary statistics of the evaluation datasets.
    }

% \vspace{-0.4cm}
\end{table}

\subsection{Baselines}

We compare the proposed \textbf{LFTC} with the following two categories of models:

\subsubsection{Non-Pre-training Models.} \textbf{TF-IDF + LR} combines TF-IDF (Term Frequency-Inverse Document Frequency) feature extraction with LR (Logistic Regression) classification algorithm. \textbf{CNNs} and \textbf{LSTM} use pre-trained GloVe word embeddings to initialize the text, which is then input into the respective deep networks. For CNNs, we compare various versions, including very deep CNNs (\textbf{VDCNN}) \cite{conneau2017very}, character CNNs (\textbf{charCNN}) \cite{zhang2015character}, recurrent CNNs (\textbf{RCNN}) \cite{lai2015recurrent}, and \textbf{textCNN}. For LSTM, we compare the \textbf{Bi-LSTM} with attention \cite{wang2016attention}. Additionally, we compare the mainstream frameworks \textbf{fastText} \cite{joulin2017bag}, Hierarchical Attention Networks (\textbf{HAN}) \cite{yang2016hierarchical}, and the lightweight model \textbf{gzip} \cite{jiang2023low}.

\subsubsection{Pre-training Models.} \textbf{BERT} \cite{devlin2019bert} is a powerful baseline model in text classification, consistently demonstrating excellent performance due to its extensive resource support. Comparing our lightweight approach with BERT highlights the advantages of our proposed method more significantly. We also compare \textbf{SentBERT} \cite{reimers2019sentence}, which fine-tunes the pre-trained BERT model to generate high-quality sentence representations tailored for specific tasks, and \textbf{mBERT} \cite{pires2019multilingual}, which handles text data in multiple languages and has cross-lingual representation capabilities. Word2Vec (\textbf{W2V}) is also considered, as it is highly useful in text classification tasks for generating word embeddings that map words into high-dimensional vector spaces, capturing semantic relationships between words.

\subsection{Implementation Details}
In our model evaluation, we conducted experiments using both the full dataset and few-shot dataset. For the kirnews, kinnews and SwahiliNews datasets, we set the shot size to 5 for the experiments. For the AGNews and SogouNews datasets, we experimented with shot sizes of 5, 10, 50 and 100. It is worth noting that when a tie occurs in $\textbf{KNN}$ (i.e., when two or more nearest neighbor labels appear with the same frequency), we only use the closest one as the final prediction. This ensures a fairer comparison of performance differences between models. This approach avoids the accuracy inflation observed in the method proposed by \cite{jiang2023low}, where they selected the two closest distances as the prediction result.

Additionally, the \textbf{LFTC} model does not require extensive computational resources and can efficiently complete text classification tasks even using only a CPU. We set the same number of threads to compare the speed with other lightweight models.

\section{Results and Analyses}
In this section, we report the results of \textbf{LFTC} on both in-distribution (ID) datasets and out-of-distribution (OOD) datasets. ID datasets refer to those where the data distribution seen during training is similar to that encountered during testing. In other words, the patterns and features learned by the model during training are also present in the test data. Conversely, OOD datasets refer to datasets where the data distribution significantly differs from the training data. Testing \textbf{LFTC} on OOD datasets helps evaluate the model's generalization ability, that is, its performance when encountering new data that differ from the training data.

\subsection{Result on ID Datasets}
Table~\ref{experiment} presents the results of \textbf{LFTC} on ID datasets. It can be observed that our method has surpassed all non pre-training models on the R8 and AGnews datasets. We also achieve competitive results on the R52 and 20News datasets. Overall, BERT-based models demonstrate strong robustness on ID datasets. However, it is known that pre-training models often require substantial data and computational resources. Our model, without any pre-training or additional data augmentation, still achieves commendable performance on ID datasets. This advancement promotes the application of parameter-free methods in text classification and inspires that efficient task processing can be achieved without relying on traditional complex model stacking. In Table~\ref{avg-table}, we present the average performance of all baseline models. It is evident that our method significantly exceeds the average on all datasets, except for the Ohsumed dataset.

\begin{table*}[t]
    \centering
    \resizebox{0.87\textwidth}{!}{
    \begin{tabular}{llcccclcccccc}
    \Xhline{1.0pt}
   &&\multicolumn{11}{c}{ Dataset } \\
    \cline { 3 - 13 }
    Category&\multirow{1}{*}{Model}&&R52&&Ohsumed&&20News & & AGnews&&R8 & \\
    % \hline
    \hline
    
    \multirow{10}{*}{Non Pre-training}
    &TFIDF+LR&&  {0.874}&&{0.549}&&{0.827} & &   {0.898}&&  {0.949} & \\
    &LSTM&&  {0.855}&&  {0.411}&&  {0.657} & &   {0.861}&&  {0.937} & \\
    &Bi-LSTM+Attn&&  {0.886}&&{0.481}&&  {0.667} & &   {0.917}&&  {0.943} & \\
    &HAN &&{0.914}&&{0.462}&&  {0.646}  & &   {0.896}&&  {0.960} & \\
    &charCNN &&  {0.724}&&  {0.269}&&  {0.401}  & &   {0.914}&&  {0.823} & \\
    &textCNN &&  {0.895}&&{0.570}&&  {0.751}  & &   {0.817}&&  {0.951} & \\
    &RCNN &&  {0.773}&&{0.472}&&  {0.716}  & &   {0.912}&&  {0.810} & \\
    &fastText &&  {0.571}&&  {0.218}&&  {0.690}  & &   {0.911}&&  {0.827} & \\
    &VDCNN &&  {0.750}&&  {0.237}&&  {0.491}  & &   {0.913}&&  {0.858} & \\
    &gzip &&  {0.852}&&  {0.365}&&  {0.608}  & &   {0.835}&&  {0.913} & \\
    
    \hdashline
    \multirow{3}{*}{Pre-training}
    &BERT &&0.960&&0.741&&0.868  & & 0.944&&0.982 &  \\

    &SentBERT &&{0.910}&&0.719&& {0.778} && {0.940}&& {0.947} &\\

    &W2V &&  {0.856}&&  {0.284}&& {0.460} && {0.892}&& {0.930} &  \\
    
    \hdashline
    
    \multirow{1}{*}{Non Pre-training}
    &\textbf{LFTC}(Ours)&&0.906&&0.435&&0.814  & & 0.919&&0.965 &\\
    
    \Xhline{1.0pt}
    \end{tabular}}
    \caption{\label{experiment}
    The accuracy of text classification by different models on the ID dataset. We test the model's performance using \textit{KNN} with \textit{K}=1.
    }
\end{table*}

\begin{table*}[t]
    \centering
    \resizebox{0.95\textwidth}{!}{
    \begin{tabular}{llcccclcccccc}
    \Xhline{1.0pt}
    && \multicolumn{11}{c}{ Dataset } \\
    \cline { 3 - 13 }
    Category&\multirow{1}{*}{Model} & \multicolumn{2}{c}{ kirnews } && \multicolumn{2}{c}{ kinnews } &&\multicolumn{2}{c}{ SwahiliNews } &&\multicolumn{2}{c}{ SogouNews }  \\
    % \hline
    
    \cline { 3 - 4 } \cline { 6 - 7 } \cline { 9 - 10 } \cline { 12 - 13 } 
    & & Full & 5-shot && Full & 5-shot && Full & 5-shot & & Full & 5-shot \\
    \hline
    
    \multirow{5}{*}{Non Pre-training}
    &Bi-LSTM+Attn &0.872& 0.254 && 0.843 & 0.253 & &0.863 &0.357 && 0.952&0.534\\
    &HAN  &0.881& 0.190 && 0.820 & 0.137  & & 0.887 &0.264 && \textbf{0.957}&0.425\\
    &fastText  &0.883& 0.245 && 0.869 & 0.170  & & 0.874 &0.347 && 0.930&0.545\\
    &gzip  &0.858&0.416  && 0.835 &0.326 && 0.850 &0.467 &&\textbf{0.957}&0.507\\
    
    \hdashline
    \multirow{3}{*}{Pre-training}
    &BERT  &0.879& 0.386 && 0.838 & 0.240  & & 0.897  & 0.396 && 0.952&0.221\\
    &W2V  &0.904& 0.288 && 0.874 & 0.281 & & 0.892  &0.373 && 0.943&0.141 \\
    &SentBERT  &0.886& 0.314 && 0.788 & 0.314  & & 0.822  &0.436&& 0.860&0.485\\
    &mBERT  &0.874& 0.324 && 0.835 & 0.229  & & 0.906  &0.558&& 0.953&0.282\\
    
    \hdashline
    \multirow{1}{*}{Non Pre-training}
    &\textbf{LFTC}(Ours) &\textbf{0.914}&\textbf{0.577}  && \textbf{0.925} & \textbf{0.391}  & & \textbf{0.916} & \textbf{0.642} && 0.935 &\textbf{0.601}\\
    
    \Xhline{1.0pt}
    \end{tabular}}
    \caption{\label{OOD}
    The accuracy of text classification by different models on the OOD dataset. We test the model's performance using \textit{KNN} with \textit{K}=1. We conduct ten experiments with 5-shot settings based on 95\% confidence and report the average accuracy, with the best performance highlighted in bold.
    }
\end{table*}

\begin{table}
\centering
\begin{tabular}{llcccclcccccc}
\Xhline{1.0pt}
&\multirow{1}{*}{Dataset} && Average && \textbf{LFTC}(Ours)\\
\hline
&R8  && 0.910 && \textbf{0.965} \\
&R52 && 0.832 && \textbf{0.906} \\
&Ohsumed && \textbf{0.445} && 0.435\\
&20News && 0.704 && \textbf{0.810}\\
&AGnews  && 0.896 && \textbf{0.919} \\
\Xhline{1.0pt}
\end{tabular}
\caption{\label{avg-table}
Comparison of \textbf{LFTC} and the average accuracy scores of all baseline models.
}
\end{table}

\begin{figure}[t]
  \includegraphics[width=\columnwidth]{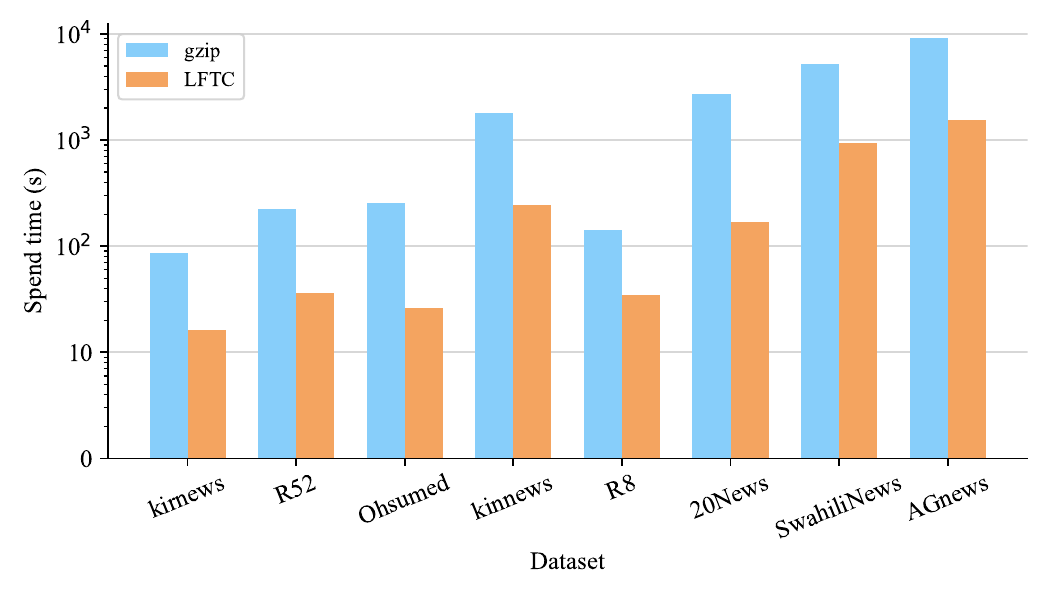}
  \caption{Time consumption of the two lightweight models across different datasets. The vertical axis is displayed in exponential form.}
  \label{fig:time01}
\end{figure}

\begin{figure}[t]
  \includegraphics[width=\columnwidth]{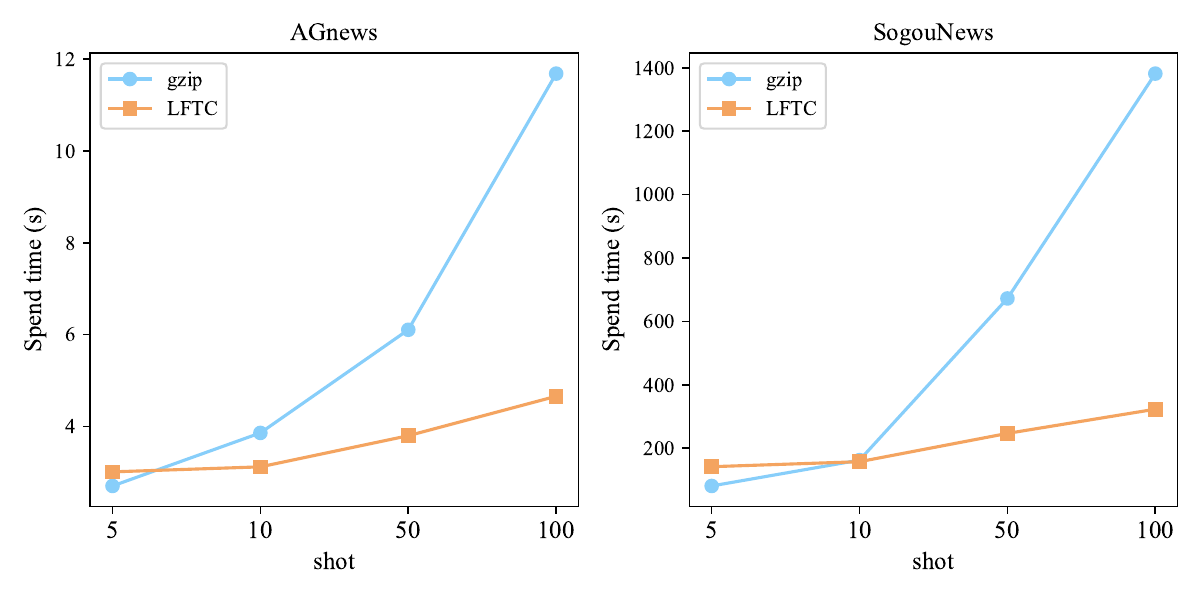}
  \caption{Comparison of time consumption of the two models in the few-shot experiments.}
  \label{fig:time02}
\end{figure}
\subsection{Result on OOD Datasets}

Table~\ref{OOD} presents the performance of \textbf{LFTC} on the OOD datasets. We observe that \textbf{LFTC} achieves SOTA results on the KirNews, KinNews, and SwahiliNews datasets, both in the full dataset and 5-shot experiments. In the full dataset experiments, \textbf{LFTC} scores 3.5\%, 8.7\%, and 1.9\% higher than BERT \cite{devlin2019bert}, respectively. Although \textbf{LFTC} performs less well on the SogouNews full dataset, it still achieves competitive scores. In the 5-shot experiments across the four datasets, \textbf{LFTC} outperforms the second-place scores by 16.1\%, 6.5\%, 8.4\%, and 5.6\%, respectively.

\subsection{Result on Few-Shot Experiment}
Due to the limited number of samples in some datasets, which prevents reaching the 100-shot sample size requirement for certain categories, we selected the larger datasets, SogouNews and AGnews, for the few-shot experiment. Tables~\ref{SogouNews} and ~\ref{AGNews} present the few-shot experiment results of \textbf{LFTC} on SogouNews and AGnews, respectively. It can be seen that on the SogouNews dataset, \textbf{LFTC} achieves state-of-the-art performance regardless of the shot settings. On the AGnews dataset, \textbf{LFTC} achieves competitive scores in the 5-shot and 10-shot experiments, although it does not surpass the SentBERT \cite{reimers2019sentence} pre-trained language model. However, in the 50-shot and 100-shot experiments, \textbf{LFTC} performed exceptionally well, achieving the best performance.

\begin{table*}[t]
    \centering
    \resizebox{0.82\textwidth}{!}{
    \begin{tabular}{llcccccccc}
    \Xhline{1.0pt}
    % \hline
    % \hline
    &\multirow{2}{*}{Model}& \multicolumn{8}{c}{ SogouNews } \\
    % \hline
    
    \cline { 3 - 3 } \cline { 5 - 5 } \cline { 7 - 7 } \cline { 9 - 9 } 
    && 5-shot && 10-shot && 50-shot && 100-shot \\
    \hline

    &Bi-LSTM+Attn &$0.534\pm\text{\scriptsize 0.042}$ && $0.614\pm\text{\scriptsize 0.047}$ &&$0.771\pm\text{\scriptsize 0.021}$ &&$0.812\pm\text{\scriptsize 0.008}$\\
    &HAN  &$0.425\pm\text{\scriptsize 0.072}$&&$0.542\pm\text{\scriptsize 0.118}$ &&$0.671\pm\text{\scriptsize 0.102}$ &&$0.808\pm\text{\scriptsize 0.020}$\\
    &fastText &$0.545\pm\text{\scriptsize 0.053}$&&$0.652\pm\text{\scriptsize 0.051}$&&$0.782\pm\text{\scriptsize 0.034}$ &&$0.809\pm\text{\scriptsize 0.012}$\\
    
    &BERT &$0.221\pm\text{\scriptsize 0.041}$&& $0.226\pm\text{\scriptsize0.060}$ &&$0.392\pm\text{\scriptsize 0.276}$&&$0.679\pm\text{\scriptsize 0.073}$\\
    &W2V  &$0.141\pm\text{\scriptsize 0.005}$&& $0.124\pm\text{\scriptsize 0.048}$ &&$0.133\pm\text{\scriptsize 0.016}$&&$0.395\pm\text{\scriptsize 0.089}$\\
    &SentBERT &$0.485\pm\text{\scriptsize0.043}$&& $0.501\pm\text{\scriptsize0.041}$ &&$0.565\pm\text{\scriptsize0.013}$ &&$0.572\pm\text{\scriptsize0.003}$\\
    &gzip  &$0.507\pm\text{\scriptsize0.042}$&&$0.574\pm\text{\scriptsize0.064}$&&
    $0.710\pm\text{\scriptsize0.010}$&&$0.759\pm\text{\scriptsize0.007}$\\
    
    \hdashline
    
    &\textbf{LFTC}(Ours) &$\textbf{0.601}\pm\text{\scriptsize0.116}$&&$\textbf{0.654}\pm\text{\scriptsize0.073}$&& $\textbf{0.807}\pm\text{\scriptsize0.022}$&&$\textbf{0.842}\pm\text{\scriptsize0.018}$&\\
    
    % \hline
    \Xhline{1.0pt}
    \end{tabular}}
    \caption{\label{SogouNews}
 Few-shot experiment on the SogouNews dataset, reporting the average accuracy of ten trials.
  }
\end{table*}

\begin{table*}[t]
    \centering
    \resizebox{0.82\textwidth}{!}{
    \begin{tabular}{llcccccccc}
    \Xhline{1.0pt}
    % \hline
    % \hline
    &\multirow{2}{*}{Model}& \multicolumn{8}{c}{ AGNews } \\
    % \hline
    \cline { 3 - 3 } \cline { 5 - 5 } \cline { 7 - 7 } \cline { 9 - 9 } 
    && 5-shot && 10-shot && 50-shot && 100-shot \\
    \hline
    
    &Bi-LSTM+Attn &$0.269\pm\text{\scriptsize 0.022}$ && $0.331\pm\text{\scriptsize 0.028}$ &&$0.549\pm\text{\scriptsize 0.028}$ &&$0.665\pm\text{\scriptsize 0.019}$\\
    &HAN  &$0.274\pm\text{\scriptsize 0.024}$&&$0.289\pm\text{\scriptsize 0.020}$ &&$0.340\pm\text{\scriptsize 0.073}$ &&$0.548\pm\text{\scriptsize 0.031}$\\
    &fastText &$0.273\pm\text{\scriptsize 0.021}$&&$0.329\pm\text{\scriptsize 0.036}$&&$0.550\pm\text{\scriptsize 0.008}$ &&$0.684\pm\text{\scriptsize 0.010}$\\
    &W2V  &$0.388\pm\text{\scriptsize 0.186}$&& $0.546\pm\text{\scriptsize 0.162}$ &&$0.531\pm\text{\scriptsize 0.272}$&&$0.395\pm\text{\scriptsize 0.089}$\\
    &SentBERT &$\textbf{0.589}\pm\text{\scriptsize0.038}$&& $\textbf{0.617}\pm\text{\scriptsize0.034}$ &&$0.706\pm\text{\scriptsize0.026}$ &&$0.713\pm\text{\scriptsize0.011}$\\
    &gzip&$0.362\pm\text{\scriptsize0.035}$&&$0.405\pm\text{\scriptsize0.060}$&&
    $0.517\pm\text{\scriptsize0.016}$&&$0.566\pm\text{\scriptsize0.022}$\\
    
    \hdashline
    
    &\textbf{LFTC}(Ours) &$0.530\pm\text{\scriptsize0.094}$&&$0.594\pm\text{\scriptsize0.102}$&& $\textbf{0.762}\pm\text{\scriptsize0.059}$&&$\textbf{0.761}\pm\text{\scriptsize0.043}$&\\
    
    % \hline
    \Xhline{1.0pt}
    \end{tabular}}%
    \caption{\label{AGNews}
    Few-shot experiment on the AGNews dataset, reporting the average accuracy of ten trials.
    }
\end{table*}

\subsection{Comparison of Model Speed}

To explore the high availability of \textbf{LFTC} in industrial production, we compared its speed with \textbf{gzip} \cite{jiang2023low}, another lightweight model, using the same parameters. Figure~\ref{fig:time01} shows the time consumption of the two models across different datasets. We observe that on the KirNews dataset, \textbf{LFTC} completes the classification task in approximately 10 seconds, whereas \textbf{gzip} requires about 10 times longer. On other datasets, \textbf{LFTC}'s runtime is faster than \textbf{gzip} by the following multiples: 6.24 times on R52, 9.74 times on KinNews, 7.29 times on Ohsumed, 4.12 times on R8, 15.80 times on 20News, 5.54 times on SwahiliNews, and 5.91 times on AGNews. Figure~\ref{fig:time02} shows the time consumption of the two models in the few-shot experiments. It can be observed that as the sample size increases, \textbf{gzip}'s time consumption increases dramatically, whereas \textbf{LFTC} does not exhibit this phenomenon.

These results demonstrate that we have successfully achieved an optimal balance between performance and resource consumption. The \textbf{LFTC} model significantly reduces computation time while maintaining high performance and low complexity. This time-saving not only enhances overall efficiency but also validates the efficiency of \textbf{LFTC} in handling large-scale text classification tasks.

\subsection{Ablation Study}
\begin{table}
    \centering
    \resizebox{0.95\columnwidth}{!}{
    \begin{tabular}{lccccccc}
    \Xhline{1.0pt}
    & \multicolumn{4}{c}{Dataset}  \\
    \cline { 2 - 5 }
    \multirow{1}{*}{Model}  &kirnews & kinnews & R8 & R52 \\
    % \hline
    \hline
    \textbf{LFTC} & \textbf{0.914} & \textbf{0.925} & \textbf{0.965} & \textbf{0.906}\\

    LFTC-MCC &0.903  &0.878 &0.933 & 0.871   \\
    LFTC-CR  & 0.883 &0.867 & 0.938 & 0.848  \\
    \Xhline{1.0pt}
    \end{tabular}}%
    \caption{\label{study}
    Ablation results of various experimental settings.
  }
\end{table}
We consider two ablation experiments on the \textbf{LFTC} model. We first remove the Multi Compressor Classification (MCC), meaning that we only use a single Zstd compressor for text compression instead of constructing multiple compressor lists for each label. The results in Table~\ref{study} show that the absence of the compressor structure leads to a noticeable decrease in performance across all datasets, with the largest drop of 4.7\% observed on the kinnews dataset. This indirectly confirms the effectiveness of our compressor structure.

The second experiment removes Centralized Reasoning (CR) from \textbf{LFTC}. In this case, we select only the result with the smallest compression length from the compressor list as the final prediction, without considering the second possible result. We observe that this leads to a significant decline in model performance, indicating that the ignored result could potentially be the correct prediction label. Based on this observation, we also attempted to consider the third similar result but did not achieve the expected scores, so we discarded this idea.

\section{Conclusion}
% In this work, we propose a text classification model based on intra-class and inter-class text information, utilizing a compressor structure to compute compression distances: \textbf{LFTC}. The model has significant value in industrial applications. Extensive experiments show that, compared to other methods, our approach requires less computational and data resources while achieving more efficient text classification within a shorter time frame, resulting in dual optimization in performance and resource usage. This approach provides an insight: rather than relying on traditional complex pre-training processes and large model structures, high-efficiency text classification can be achieved through innovative compressor structure design and utilization of valuable information. Such a strategy not only enhances the practical applicability of the model but also offers a new perspective for machine learning tasks in resource-constrained environments.

In this work, we propose a text classification model \textbf{LFTC} based on the compressor structure which computes compression distances through intra-class and inter-class text information. Extensive experiments show that, compared to other methods, our method requires less computational and data resources while achieving more efficient text classification within a shorter time frame, resulting in dual optimization in performance and resource usage. This method provides an insight: rather than relying on traditional complex pre-training processes and large model structures, high-efficiency text classification can be achieved through innovative compressor structure design and utilization of valuable information. Such a strategy not only enhances the practical applicability of the model but also offers a new perspective for machine learning tasks in resource-constrained environments.


\section*{Limitations}
\textbf{LFTC} emphasizes dual optimization of both speed and performance for text classification tasks, and we have not pursued extreme performance optimization at the expense of reduced speed. For example, when constructing the compressor list, we considered that having too many compressors in the list could affect the model’s speed, so we limited the number of compressors in the list. This approach limits our performance scores in some experiments. Another limitation of \textbf{LFTC} is that we adjusted the compression levels according to different datasets, but we did not adjust the compression levels for each individual data point within the datasets. We speculate that more targeted adjustments of compression levels for specific data points could obtain better performance scores.

\section*{Ethics Statement}
Our proposed \textbf{LFTC} demonstrates outstanding advantages and is an excellent solution for text classification tasks. This method is only evaluated on publicly available datasets to ensure that personal privacy is not compromised. In addition, we also provide the source code implementation of \textbf{LFTC}, enabling researchers to realistically reproduce its performance and promote academic exchange in the field of text classification.
\bibliography{custom}

\appendix

% \section{Appendix}
% \label{sec:appendix}
\newpage
\section{Execution Process}
The pseudocode for the \textbf{LFTC} model execution process is shown in Algorithm~\ref{alg:lftc_model}.
\begin{algorithm}
\caption{The execution process of \textbf{LFTC}}
\label{alg:lftc_model}
\textbf{Compressor List Construction}

\begin{algorithmic}[1]
\STATE \textbf{Input:} Texts from each class $C_i$, step size $S$.
\STATE \textbf{Output:} Compressor set $Z_{C_i}$ for each class.
\FOR{each category $C_i$}
    \STATE Divide texts $Ts_{C_i}$ into $N_{C_i}$ blocks.
    \STATE \text{For each block:}
    \FOR{$x \gets 0$ \textbf{to} $N_{C_i}-1$}
        \STATE Build compressor:
        \STATE$Z_{C_i, N_{C_i}} = $\\$Zstd\left(Ts_{C_i}[x \cdot S:(x+1) \cdot S]\right)$
    \ENDFOR
    \STATE Save compressors list:
    \STATE $Z_{C_i} = \{Z_{C_i, 0}, Z_{C_i, 1}, \ldots, Z_{C_i, N_{C_i}-1}\}$.
\ENDFOR
\STATE \textbf{Return} Compressor set $Z_{C_i}$ for each class.
\end{algorithmic}
\textbf{Multi Compressor Classification}

\begin{algorithmic}[1]
\STATE \textbf{Input:} Prediction text $T_{C_i}$.
\STATE \textbf{Output:} The two labels with the lowest scores,  $p$ and $q$.
\STATE Calculate $\text { L }_{\max }$ using Eq.4.
\STATE Compressed substring\\ $L_{\text {d}} = T_{C_i}:\operatorname{Re}(\text { L }_{\max }, p)$.
\STATE Entropy encoding: \\$\mathcal{L}e = -\sum_{i=1}^n P_{(L_d)} \cdot \log_2{P_{(L_d)}}$.
\STATE $Score(Z_{C_i, N_{C_i}}) = \mathcal{L}e(T_{C_i}) + O(\mathcal{D})$.
\STATE $p,q=min(Score)$.
\STATE \textbf{Return} $p$ and $q$.
\end{algorithmic}

\textbf{Centralized Reasoning}
\begin{algorithmic}[1]
\STATE \textbf{Input:} Text labeled as $p$ and $q$, test text $T_{C_i}$.
\STATE \textbf{Output:} Predicted label for text $T_{C_i}$.
\STATE Extracts text with labels $p$ and $q$ from training data.
\STATE Excludes $T_p $ and $T_q $, Obtain Gold data.
\STATE Calculate NCD between $T_{C_i}$ and Gold data using Eq.8.
\STATE Use $KNN$ to determine the classification of $T_{C_i}$.
\STATE \textbf{Return} Predicted label for text $T_{C_i}$.
\end{algorithmic}
\end{algorithm}

\section{Detail Display}
\begin{figure}[ht]
    \centering
    \begin{minipage}{\columnwidth}
        \centering
        \includegraphics[width=\textwidth]{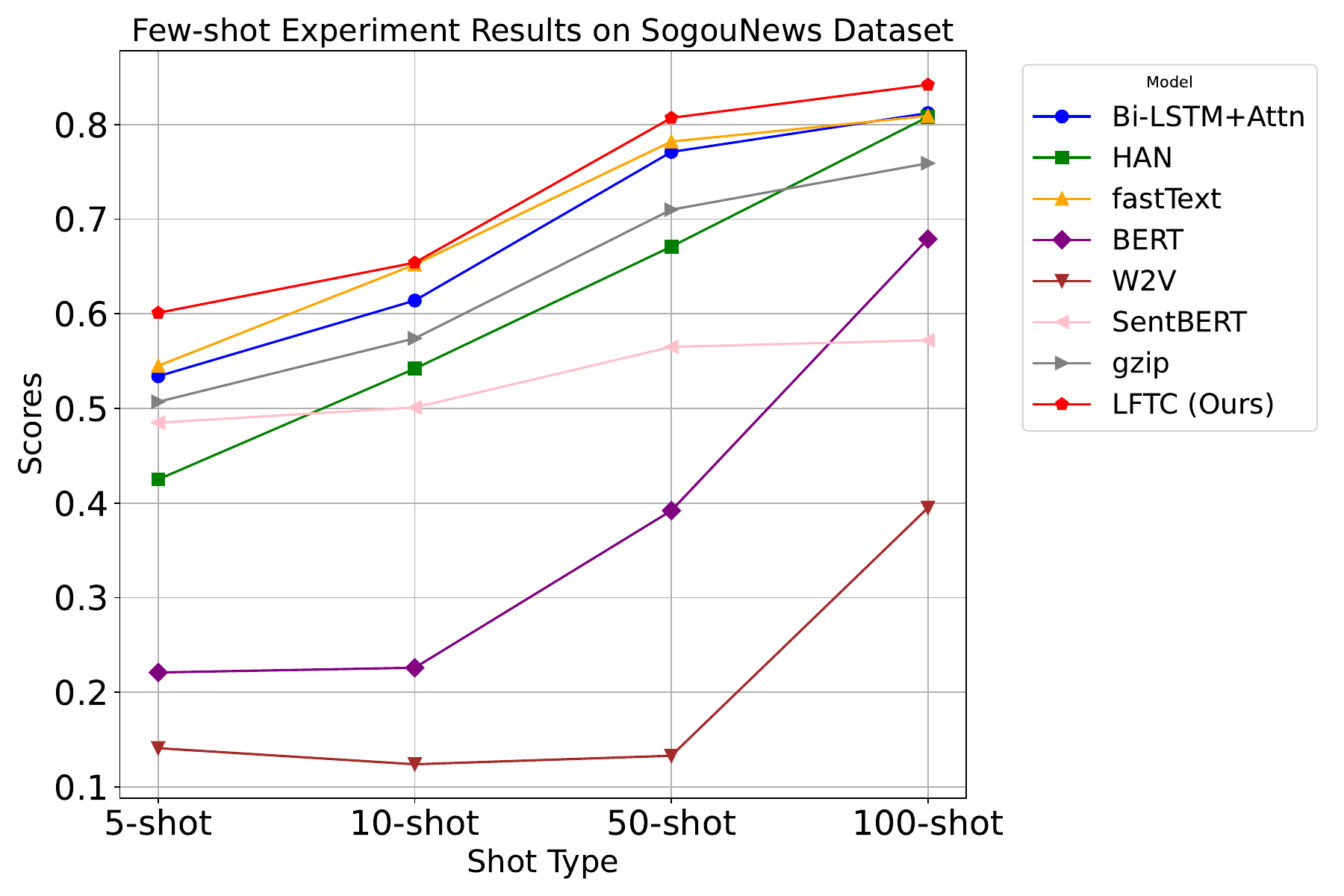}
        \caption{Comparison of Few-shot experimental performance between different methods on SogouNews.}
        \label{fig:image1}
    \end{minipage}\hfill
    \begin{minipage}{\columnwidth}
        \centering
        \includegraphics[width=\textwidth]{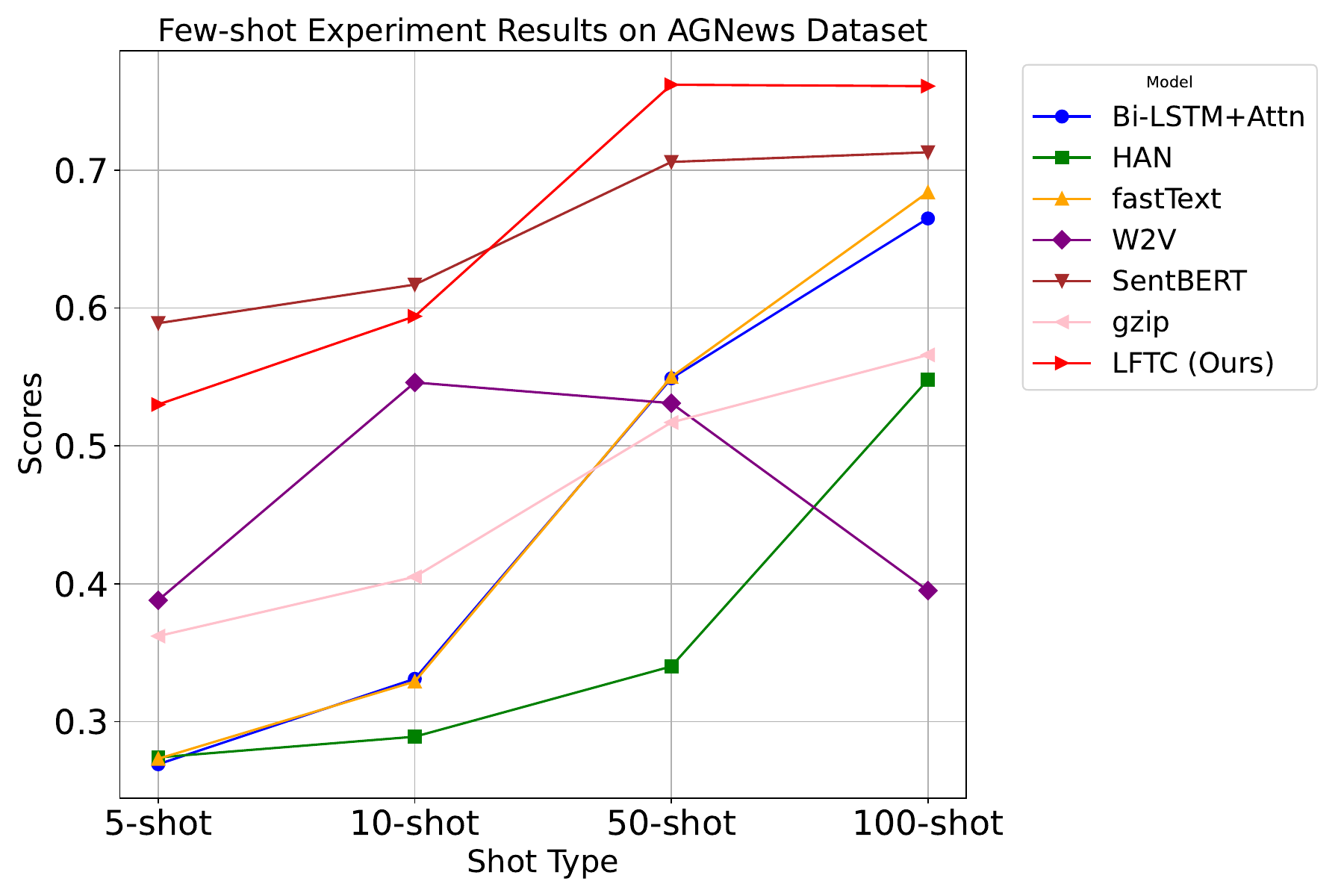}
        \caption{Comparison of Few-shot experimental performance between different methods on AGNews.}
        \label{fig:image2}
    \end{minipage}
\end{figure}

Figures~\ref{fig:image1} and \ref{fig:image2} respectively show the performance scores of different models in the Few shot experiment on two datasets. It can be seen that \textbf{LFTC} has achieved good results.

Table~\ref{spent_times}, \ref{tab:time_comparison}, and \ref{tab:time_comparison_sogou} provide detailed information on the time required for the model in different experiments. We conducted ten experiments and took the average. We can observe that compared to \textbf{gzip} \cite{jiang2023low}, which is also a lightweight model, we greatly reduce the time consumption.

\begin{table*}[t]
    \centering
    \resizebox{0.95\textwidth}{!}{
    {\fontsize{10pt}{12pt}\selectfont
    \begin{tabular}{lcccccccc}
    \Xhline{1.0pt}
    % \hline
    Dataset & kirnews & R52 & Ohsumed & kinnews & R8 & 20News & SwahiliNews & AGNews \\
    \hline
    gzip\_spent & 85.74 & 225.66 & 252.74 & 1794.90 & 143.37 & 2690.40 & 5243.49 & 9085.00 \\
    LFTC\_spent & 16.08 & 36.17 & 25.95 & 246.18 & 34.78 & 170.27 & 946.16 & 1537.32 \\
    % \hline

    \Xhline{1.0pt}
    \end{tabular}
    }}
    \caption{\label{spent_times}
    Comparison of the time spent (in seconds) by gzip \cite{jiang2023low} and LFTC across various datasets.
    }
\end{table*}

\begin{table}[ht]
    \centering
    {\fontsize{10pt}{12pt}\selectfont
    \begin{tabular}{lcc}
        \Xhline{1.0pt}
        AGNews & gzip Time & LFTC Time \\
        \hline
        5-shot  & 2.70 & 3.01 \\
        10-shot & 3.85 & 3.12 \\
        50-shot & 6.10 & 3.79 \\
        100-shot & 11.68 & 4.65 \\
        \Xhline{1.0pt}
    \end{tabular}}
    \caption{Comparison of time spent by gzip and LFTC models in AGNews Few-shot experiment.}
    \label{tab:time_comparison}
\end{table}

\begin{table}[ht]
    \centering
    % \resizebox{0.60\columnwidth}{!}{
    {\fontsize{10pt}{12pt}\selectfont
    \begin{tabular}{lcc}
        \Xhline{1.0pt}
        SogouNews & gzip Time & LFTC Time \\
        \hline
        5-shot  & 80.84 & 141.61 \\
        10-shot & 162.40 & 157.40 \\
        50-shot & 672.41 & 246.67 \\
        100-shot & 1381.72 & 322.99 \\
        \Xhline{1.0pt}
    \end{tabular}}
    % }
    \caption{Comparison of time spent by gzip and LFTC models in SogouNews Few-shot experiment.}
    \label{tab:time_comparison_sogou}
\end{table}

\section{Experiment Replication}
In this appendix, we provide the process for reproducing the experimental results.

To reproduce the experimental results from the paper ``Low-Resource Fast Text Classification Based on Intra-Class and Inter-Class Distance Calculation", you should first execute the `main\_text.py' file located in the root directory of the code. By default, the experiments will conduct text classification on the `kirnews' dataset. If you wish to experiment with other datasets, you can change the `dataset' parameter in the code to specify the desired dataset name. For example, if you want to use the `R8' dataset, simply modify the `dataset' parameter to `R8'.

For datasets downloaded in parquet format from Huggingface, you will need to convert them to CSV format using the simple preprocessing script `parquet\_to\_csv.py' in the root directory of the code. After conversion, you can directly use the converted CSV dataset for the experiments.

To conduct Few-Shot experiments, set the `all\_train' parameter to `False' and the `num\_train' parameter to the desired number of Few-Shot samples. This will allow you to train and evaluate the model with a limited number of samples.

Make sure to standardize the data before running the experiments and to thoroughly document the experimental configuration and process to ensure reproducibility and reliability of the results. By maintaining a modular structure and detailed documentation, the experiments can be made more maintainable and scalable for future work.

\section{Advantages of LFTC}
With the continuous development of society, the amount of information across various fields is experiencing exponential growth. We not only need to constantly improve the performance of text classification models but also must focus on their processing speed and generalization capabilities \cite{kowsari2019text}.

We have designed a unique compressor structure for \textbf{LFTC}, which maximizes the utilization of intra-class regularity information to achieve efficient classification tasks. Additionally, we have minimized the inclusion of redundant data irrelevant to classification, relying solely on inter-class information from Gold data to obtain the final prediction. These two modules not only enhance the performance of the original lightweight classification model but also significantly reduce processing time. It can be said that \textbf{LFTC} provides a dual optimization solution for text classification tasks.

The \textbf{LFTC} model has demonstrated outstanding performance across multiple text classification datasets, particularly in minority language classification tasks such as kinnews and kirnews, where it has achieved results surpassing those of large pre-trained language models like BERT. This further proves \textbf{LFTC}'s high generalization ability.

\section{Future Work Discussion}
AI tasks typically rely on algorithm optimization and resource investment. On one hand, it is necessary to continuously improve algorithms and model architectures to enhance performance. On the other hand, high-quality data and powerful computational resources are also essential. The \textbf{LFTC} model, as a parameter-free text classification model, surpasses the BERT model, which has a large number of parameters, to a certain extent. This achievement suggests that, while optimizing algorithms and model architectures, we can also effectively mitigate resource constraints, which is particularly important in the era of large-scale language models (LLMs).

In the future, we plan to extend the compressor architecture from \textbf{LFTC} to the image classification domain. Recent research indicates that existing neural network compressors and combinations based on compressor distance metrics can outperform traditional models in image classification tasks \cite{jiang2022few}. We believe that by applying the compression technology from \textbf{LFTC} to image classification, we can further improve model performance while reducing computational resource requirements.

\end{document}